\pdfoutput=1
\documentclass{article}

\PassOptionsToPackage{numbers}{natbib}
\usepackage[main,final]{neurips_2026}
\usepackage{microtype}
\usepackage{graphicx}
\usepackage{subcaption}
\usepackage{booktabs} %
\usepackage{multirow}
\usepackage[table]{xcolor}
\usepackage{xspace}
\usepackage[ruled,vlined]{algorithm2e}
\usepackage{algorithmic}

\usepackage{hyperref}

\usepackage{amsmath}
\usepackage{amssymb}
\usepackage{mathtools}
\usepackage{amsthm}

\usepackage[capitalize,noabbrev]{cleveref}

\theoremstyle{plain}

\theoremstyle{definition}

\theoremstyle{remark}

\newcommand{\vQ}{\mathbf{Q}}
\newcommand{\vK}{\mathbf{K}}
\newcommand{\vV}{\mathbf{V}}
\newcommand{\vdQ}{\mathbf{dQ}}
\newcommand{\vdK}{\mathbf{dK}}
\newcommand{\vdV}{\mathbf{dV}}
\newcommand{\vS}{\mathbf{S}}
\newcommand{\vdS}{\mathbf{dS}}
\newcommand{\vP}{\mathbf{P}}
\newcommand{\vdP}{\mathbf{dP}}

\newcommand{\vO}{\mathbf{O}}
\newcommand{\vdO}{\mathbf{dO}}

\newcommand{\vH}{\mathbf{H}}
\newcommand{\vZ}{\mathbf{Z}}
\newcommand{\vdH}{\mathbf{dH}}
\newcommand{\vdZ}{\mathbf{dZ}}
\newcommand{\vD}{\mathbf{D}}

\definecolor{deepgreen}{rgb}{0.0, 0.5, 0.0}  
\definecolor{deepred}{rgb}{0.6, 0.0, 0.0}  
\definecolor{darkgreen}{rgb}{0.15, 0.75, 0.15}
\definecolor{cvprblue}{rgb}{0.21,0.49,0.74}
\definecolor{lightblue}{rgb}{0.90, 0.95, 0.99}

\newcommand{\annotate}[1]{\textcolor{gray}{{#1}}\xspace}
\newcommand*\scircled[1]{\tikz[baseline=(char.base)]{\node[shape=circle,fill,inner sep=0.3pt] (char) {\textcolor{white}{#1}};}}

\newcommand{\add}[1]{{\color{blue} #1}}

\usepackage[disable,textsize=tiny]{todonotes}

\newcommand{\our}{\texttt{SLA2}\xspace}

\usepackage[utf8]{inputenc} %
\usepackage[T1]{fontenc}    %
\usepackage{url}            %
\usepackage{amsfonts}       %
\usepackage{nicefrac}       %

\title{SLA2: Sparse-Linear Attention with Learnable Routing and Quantization-aware Training}

\author{%
  \textbf{Jintao Zhang}$^{1}$\quad \textbf{Haoxu Wang}$^{1}$\quad \textbf{Kai Jiang}$^{1}$\quad \textbf{Kaiwen Zheng}$^{1}$\quad \textbf{Youhe Jiang}$^{1}$\\
  \textbf{Ion Stoica}$^{2}$\quad \textbf{Jianfei Chen}$^{1}$\thanks{Corresponding author: \texttt{jianfeic@tsinghua.edu.cn}.}\quad \textbf{Jun Zhu}$^{1}$\quad \textbf{Joseph E. Gonzalez}$^{2}$\\
  $^{1}$Tsinghua University\quad $^{2}$UC Berkeley
}

\hypersetup{hidelinks, pdftitle={SLA2: Sparse-Linear Attention with Learnable Routing and Quantization-aware Training}, pdfauthor={Jintao Zhang, Haoxu Wang, Kai Jiang, Kaiwen Zheng, Youhe Jiang, Ion Stoica, Jianfei Chen, Jun Zhu, Joseph E. Gonzalez}}

\begin{document}

\maketitle

\begin{abstract}
 Sparse-Linear Attention (SLA) combines sparse and linear attention to accelerate diffusion models and has shown strong performance in video generation. However, (i) SLA relies on a heuristic split that assigns computations to the sparse or linear branch based on attention-weight magnitude, which can be suboptimal. Additionally, (ii) after formally analyzing the attention error in SLA, we identify a mismatch between SLA and a direct decomposition into sparse and linear attention. We propose SLA2, which introduces (I) a learnable router that dynamically selects whether each attention computation should use sparse or linear attention, (II) a more faithful and direct sparse-linear attention formulation that uses a learnable ratio to combine the sparse and linear attention branches, and (III) a sparse + low-bit attention design, where low-bit attention is introduced via quantization-aware fine-tuning to reduce quantization error. Experiments show that on video diffusion models, SLA2 can achieve 97\% attention sparsity and deliver an 18.6$\times$ attention speedup while preserving quality.
\end{abstract}

\section{Introduction}

Trainable sparse attention methods~\cite{zhang2025sla,zhang2025vsa,wu2025vmoba,zhan2025bidirectional} have shown strong performance in diffusion models. They often achieve higher attention sparsity than training-free sparse attention methods~\cite{zhangspargeattention,xi2025sparse,chen2025sparse}. Among them, Sparse-Linear Attention (SLA)~\cite{zhang2025sla} is a promising approach that introduces a linear-attention branch to compensate for the sparse-attention branch, improving overall sparsity. SLA has been validated on both image and video diffusion models, such as TurboDiffusion~\cite{zhang2025turbodiffusion}.

\textbf{Motivation of SLA.} SLA finds that, in diffusion models, the attention map $P$ could be decomposed into a high-sparse part $P_1$ and a low-rank part $P_2$, and $P = P_1 + P_2$. SLA computes $O = O_s + {\rm proj}(O_l)$, where $O_s$ and $O_l$ are the sparse- and linear-attention outputs, and ${\rm proj}$ is a trainable projection.

\textbf{Limitation of SLA and motivation of SLA2.} \textbf{(L1)} \textit{Mismatch between SLA output and the original sparse-linear decomposition.} After an analysis of the difference of the SLA formulation with the original SLA motivation, we find that the sparse attention map $P_s$ of SLA differs from the decomposed sparse attention map $P_1$ by a constant scaling factor. Specifically, we find $P_1= \alpha P_s$, where $\alpha$ is a ratio vector.
To compensate for the mismatch, SLA introduces and trains an additional linear attention projection, which may fail to fully address it. 
We therefore aim to propose a sparse-linear attention formulation that more directly matches the original motivation.
\textbf{(L2)} \textit{Heuristic routing for sparse and linear attention branches.} SLA does not optimally address the key design choice of how to split computation between the sparse and linear branches. In practice, SLA assigns attention associated with larger attention weights to the sparse branch and routes the remaining computation to the linear branch. 
This heuristic split is not optimal. For example, moving some weights from $P_1$ to $P_2$ via brute-force selection may not increase the rank of $P_2$, while still improving the sparsity of $P_1$.
We therefore aim to design a more principled split, guided by a clear optimization objective. 
Finally, low-bit attention can be introduced to SLA to obtain an additional speedup. We thus aim to incorporate low-bit attention into SLA in a way that introduces as little quantization error as possible, enabling further attention speedup.

\textbf{Our method.} We propose \our, a correct sparse-linear attention method that 
reformulates sparse linear attention to
(1) better match the original motivation, and (2) optimally route between the sparse and linear attention branches.
To address \textbf{(L1)}, we directly learn
the ratio $\alpha$ to combine the sparse and linear attention branches. This formulation aligns exactly with the sparse and linear components decomposition of attention.
To address \textbf{(L2)}, we formulate the approximation error of combining sparse attention and linear attention relative to full attention, and build a learnable sparse-attention mask predictor $\mathcal{R}$ that supports gradient backpropagation. We train this predictor by minimizing the formulated error.
Furthermore, we build low-bit attention on top of sparse attention to achieve additional attention speedups. To reduce the error introduced by low-bit quantization, we integrate the quantization process into training in a quantization-aware manner, enabling the model to better adapt to low-bit quantization and thus improve the accuracy of low-bit attention at inference time.

\textbf{Result.} \our achieves $97\%$ attention sparsity and an $18.6\times$ attention-kernel speedup on an RTX 5090. Please note that $97\%$ sparsity corresponds to about $96.5\%$--$96.8\%$ attention FLOPs savings after accounting for the linear-attention branch in \our. At $97\%$ sparsity, \our achieves higher AQ than all sparse baselines at $90\%$ sparsity on both models, although the comparison varies across metrics.

\textbf{Contribution.} Our contributions are as follows:

(1) We carefully analyze the limitations of SLA and propose \our, a more reasonable sparse-linear attention method. \our includes a learnable router that splits computation between the sparse and linear attention branches, along with a simple yet effective learnable combination for sparse and linear attention branches. 
For some insight on the design of \our, please see Sections~\ref{sec:rethink} and~\ref{sec:insight}.

(2) We integrate quantization-aware training (QAT) into \our to further accelerate attention without degrading end-to-end video generation quality, demonstrating the effectiveness of QAT for attention.

(3) Experiments show that \our achieves $97\%$ attention sparsity and an $18.6\times$ attention runtime speedup on video diffusion models while maintaining video quality, surpassing baseline methods.

\section{Preliminaries}

\subsection{Sparse-Linear Attention}
SLA (Sparse-Linear Attention)~\cite{zhang2025sla} combines sparse softmax attention and linear attention using a heuristic sparse attention mask. Below, we describe the computation of SLA.

\paragraph{Notation.}
Let $Q,K,V\in\mathbb{R}^{N\times d}$ be the query, key, and value matrices, where $N$ is the sequence length and $d$ is the head dimension. Let
\[
S={QK^\top}/{\sqrt{d}}\in\mathbb{R}^{N\times N}
\]
be the attention score matrix. We use $\mathrm{softmax}(\cdot)$ to denote row-wise softmax. We use $\phi(\cdot)$ as the activation function for linear attention.

\textbf{Mask construction.}
SLA first computes compressed attention weights using pooled queries and keys:
\begin{equation}
P_c=\mathrm{softmax}\!\left({\mathrm{pool}(Q)\,\mathrm{pool}(K)^\top}/{\sqrt{d}}\right),
\end{equation}
where $\mathrm{pool}(\cdot)$ applies mean pooling over the token dimension within each token block. For each row of $P_c$, SLA assigns the top $k_h\%$ entries to sparse attention and the bottom $k_l\%$ entries to skipping, with the remaining entries handled by linear attention. In practice, $k_l$ is typically small and can be omitted. This procedure yields a binary mask $M_c\in\{0,1\}^{N/b_q\times N/b_k}$, where the top $k_h\%$ entries in each row are set to $1$ and the others to $0$. Then, we obtain a $M \in\{0,1\}^{N \times N}$ by expanding $M_c$.

\textbf{Sparse attention output.}
Given $M$, SLA computes sparse softmax attention only on entries selected by the mask:
\begin{equation}
P_s=\mathrm{softmax}_{M}(S)\in\mathbb{R}^{N\times N}, ~ O_s = P_sV \in\mathbb{R}^{N\times d}
\end{equation}
where $\mathrm{softmax}_{M}(S)_{ij}=M_{ij}e^{S_{ij}}/\sum_r M_{ir}e^{S_{ir}}$ normalizes only the retained entries; each row retains at least one entry. We use $\odot$ for element-wise multiplication.

\textbf{Linear attention output.}
For the remaining entries ($1-M$), SLA applies linear attention:
\begin{equation}
O_l=\frac{\bigl[\phi(Q)\phi(K)^\top\odot(1-M)\bigr]V}{\bigl[\phi(Q)\phi(K)^\top\odot(1-M)\bigr]\mathbf{1}}
\in\mathbb{R}^{N\times d},
\end{equation}
where $\mathbf{1}\in\mathbb{R}^{N\times 1}$ is an all-ones vector, and the division is element-wise to perform row-wise normalization.

\textbf{Final output.}
The final SLA output is
\begin{equation}
\label{eq:sla_output}
O = O_s + \mathrm{Proj}(O_l),
\end{equation}
where $\mathrm{proj}(\cdot) \in\mathbb{R}^{d\times d}$ is a learnable linear projection.

\subsection{Rethinking Sparse-Linear Attention}  \label{sec:rethink}

\paragraph{Original motivation of Sparse-Linear Attention.}
Let $P=\mathrm{softmax}(S)\in\mathbb{R}^{N\times N}$ be the full-attention probability matrix. Given a binary mask $M\in\{0,1\}^{N\times N}$, we decompose full attention into two parts:
\begin{equation}
P = P_{1}+P_{2}, \qquad
P_{1}=P\odot M,\quad P_{2}=P\odot(1-M),
\label{eq:pf_split}
\end{equation}
where $P_{1}$ corresponds to the mask-selected attention positions (computed by sparse softmax attention), and $P_{2}$ corresponds to the remaining positions (approximated by linear attention). The motivation of SLA is to approximate $P_{1}$ with a sparse-attention distribution and approximate $P_{2}$ with a linear-attention distribution. With $V\in\mathbb{R}^{N\times d}$, the full-attention output is
\begin{equation}
O_f = PV = P_{1}V + P_{2}V ~\in~ \mathbb{R}^{N\times d}.
\label{eq:of_split}
\end{equation}

\paragraph{Error of the sparse attention branch.}
Sparse attention does not directly produce $P_{1}$, because it renormalizes probabilities over the masked positions in each row. Let $\alpha$ denote the probability sum on the masked positions for each query:
\begin{equation}
\alpha = P_{1}\mathbf{1} ~\in~ \mathbb{R}^{N\times 1},
\label{eq:alpha_def}
\end{equation}
where $\mathbf{1}\in\mathbb{R}^{N\times 1}$ is an all-one vector. The sparse-attention distribution is
\begin{equation}
P_s = \frac{P_{1}}{\alpha}\in\mathbb{R}^{N\times N},
\label{eq:ps_def}
\end{equation}
Therefore, $P_s$ is not aligned with $P_{1}$; it is obtained by row-wise normalizing $P_{1}$ so that each row sums to $1$.
In terms of attention output, with $O_s=P_sV\in\mathbb{R}^{N\times d}$, the desired sparse attention output is
\begin{equation}
P_{1}V = (\alpha\odot P_s)V = \alpha\odot O_s.
\label{eq:output_scale_mismatch}
\end{equation}
As a result, each row has a scale mismatch controlled by $\alpha$.

\paragraph{How SLA compensates for the mismatch.}
SLA output is shown in Equation~\ref{eq:sla_output}.
Comparing Equation~\ref{eq:of_split} and using Equation~\ref{eq:output_scale_mismatch}, we can interpret $\mathrm{proj}(O_l)$ as jointly accounting for the linear component $P_{2}V$ and the residual induced by the sparse attention branch mismatch:
\begin{equation}
\mathrm{proj}(O_l)\ \approx\ P_{2}V\ +\ (\alpha-\mathbf{1})\odot O_s.
\label{eq:proj_compensate}
\end{equation}
However, this correction is not directly aligned with the original decomposition motivation: the linear attention branch is also forced to offset the sparse attention branch's scaling error, making the compensation harder to learn.

\paragraph{A more reasonable formulation.}
A more faithful way to match the decomposition in Equation~\ref{eq:pf_split} is
\begin{equation}
P \ \approx\ \alpha \odot P_s\ +\ (1-\alpha)\odot P_l,
\label{eq:new_formulation_prob}
\end{equation}
where $\alpha\in\mathbb{R}^{N\times 1}$. Here, $P_s,P_l\in\mathbb{R}^{N\times N}$ are the attention-weight matrices corresponding to the sparse attention and the linear attention branches, and each is row-normalized so that every row sums to $1$. The attention output is
\begin{equation}
O \ =\ \alpha\odot (P_sV)\ +\ (1-\alpha)\odot (P_lV).
\label{eq:new_formulation_out}
\end{equation}
Here, $\alpha\odot P_s$ better matches $P_{1}$, which removes the row-wise scaling mismatch in the sparse attention branch; therefore, an extra $\mathrm{proj}(\cdot)$ on $O_l$ for compensation is no longer needed. Moreover, $(1-\alpha)$ ensures that $\alpha\odot P_s + (1-\alpha)\odot P_l$ is row-normalized, avoiding magnitude drift of the output.

\begin{figure*}[t]
    \centering

    \begin{minipage}[t]{0.45\textwidth}
        \centering
        \vspace{0pt}
        \includegraphics[width=\linewidth]{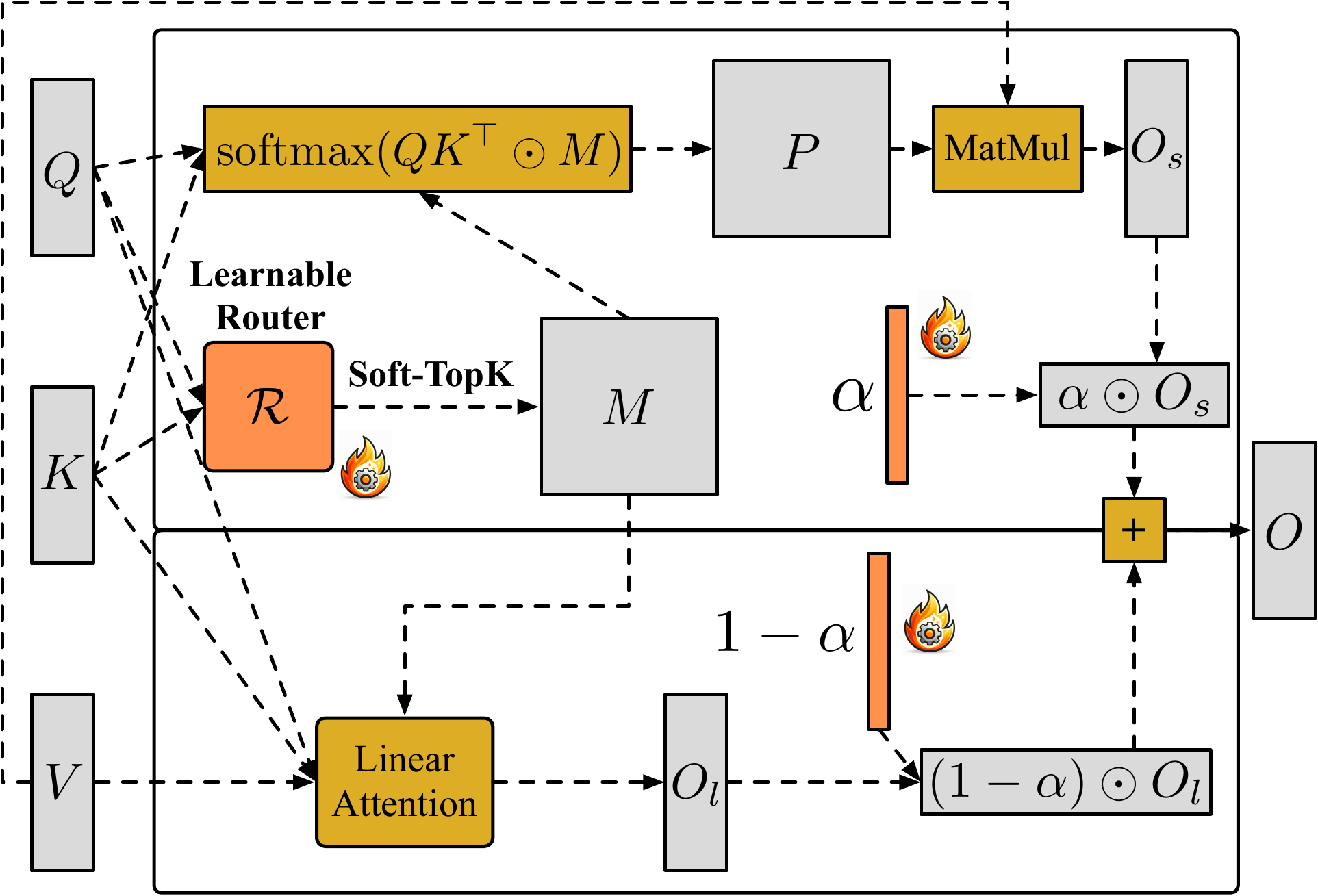}
        \vspace{-1em}
        \caption{Computation pipeline of \our.}
        \label{fig:pipeline}
    \end{minipage}
    \hfill
    \begin{minipage}[t]{0.54\textwidth}
        \centering
        \vspace{0pt}

        \refstepcounter{algocf}
        \label{alg:training}

        \hrule height 0.8pt
        \vspace{0.4em}

        \hspace{-4.5em}\noindent\textbf{Algorithm~\thealgocf: Fine-tuning using \our.}

        \vspace{0.15em}
        \hrule height 0.4pt
        \vspace{0.15em}

        \small
        \begin{algorithmic}[1]
            \STATE \textbf{Stage 1: Initialize $\mathcal{R}$ and $\alpha$:}

            \STATE Sample $Q, K, V$ tensors as dataset $D$.

            \STATE {\small ${L}$ = ${\rm MSE}({\rm Attn}(Q, K, V), \our(Q, K, V, k\%, \mathcal{R}, \alpha))$;}

            \STATE Train $\mathcal{R}, \alpha$ under different $k$\% according to the $L$;

            \STATE \textbf{Stage 2: Fine-tune the diffusion model $\Theta$ and $\alpha$:}

            \STATE Replace the attention in $\Theta$ by \our;

            \STATE Fine-tune $\Theta$, $\alpha$ using an end-to-end diffusion loss.
        \end{algorithmic}

        \vspace{0.4em}
        \hrule height 0.8pt
    \end{minipage}

    \vspace{-1em}
\end{figure*}

\section{\our Design}

Figure~\ref{fig:pipeline} illustrates the computation pipeline. According to the analysis in Section~\ref{sec:rethink} and Equation~\ref{eq:new_formulation_out}, we present the overall formulation of \our as follows:
\begin{align}
\label{equ:sla2}
O \ =\ \alpha\odot O_s\ +\ (1-\alpha)\odot O_l,
\end{align}
where $\alpha \in \mathbb{R}^{N \times 1}$ has learnable values between 0 and 1, shared within each query block (Algorithm~\ref{alg:fwd}), and
\begin{align}
\label{equ:sla2_notation}
O_s = \mathrm{softmax}_{M} (Q K^\top / \sqrt{d}) V, ~~ O_l = \mathrm{norm} (\phi(Q) \phi(K)^\top \odot (1-M)) V, ~~M = \mathcal{R}(Q, K),
\end{align}
where $\mathcal{R}$ is a learnable module, which will be explained in Section~\ref{sec:learnable_router}. $\phi(\cdot)$ is an activation function for linear attention, and we use the softmax function. $\rm{norm}$ normalizes the sum of rows to 1.

\textbf{Implementation of getting $O_s$ and $O_l$.} From Equation~\ref{equ:sla2_notation}, it may appear that computing $O_s$ and $O_l$ requires full matmuls $QK^\top$ and $PV$. In contrast, our implementation is highly efficient. For $O_s$, built on top of the FlashAttention Algorithm, we only perform the matmuls $QK^\top$ and $PV$ for the positions where $M=1$, and skip the other computations. For $O_l$, we also do not compute the matmul $QK^\top$ directly, but first compute $K^\top V$ according to the positions where $M=0$. Then we multiply $Q$ with the result. See Algorithm~\ref{alg:fwd} for more details.

\section{Learnable Router}  \label{sec:learnable_router}

The learnable router $\mathcal{R}$ aims to dynamically output a mask $M$ to decide which probabilities in $P$ should be computed by the sparse attention branch. Its decisions mainly depend on $Q$ and $K$, and are independent of $V$. We therefore take $Q$ and $K$ as inputs to $\mathcal{R}$. However, the sequence length $N$ can be large, making $\mathcal{R}$ expensive. To reduce its computational cost, we leverage the fact that adjacent tokens in $Q$ and $K$ often exhibit similar distributions~\cite{zhangspargeattention}. Following~\cite{jiang2024minference,zhangspargeattention,gao2024seerattention}, we apply mean pooling over consecutive $b_q$ and $b_k$ tokens to compress $Q$ and $K$:
\begin{equation}
\bar Q=\mathrm{pool}(Q) \in \mathbb{R}^{N/b_q\times d},~~~ \bar K=\mathrm{pool}(K)  \in \mathbb{R}^{N/b_k\times d}.
\end{equation}
To make $\mathcal{R}$ learnable, we further introduce two linear projections $\mathrm{proj}_q,\mathrm{proj}_k\in\mathbb{R}^{d\times d}$ for $\bar Q$ and $\bar K$, respectively. To get $M$, we perform
\begin{align} \label{equ:mask_of_sla2}
P_c &= \mathrm{softmax}\!\left(\mathrm{proj}_q(\bar Q)\,\mathrm{proj}_k(\bar K)^\top/\sqrt{d}\right), \notag\\
M_c &= \mathrm{Top}\text{-}\mathrm{k}\bigl(k\%, P_c\bigr)\in\mathbb{R}^{{N}/{b_q}\times {N}/{b_k}},
\end{align}
where $\mathrm{Top}\text{-}\rm{k}$ is applied row-wise, setting the top $k$\% positions to $1$ and the others to $0$. The compressed mask $M_c$ can be expanded to an $N\times N$ mask to support the computation in Equation~\ref{equ:sla2_notation}. In practice, our forward and backward GPU kernels for \our only require $M_c$, since we implement the method efficiently on top of a block-wise FlashAttention-style algorithm. We will elaborate on this in Section~\ref{sec:inference}.

Finally, we note that Top-k avoids gradient propagation during training. We therefore replace Top-k with a learnable version during training. The details and the full training procedure are provided in Section~\ref{sec:training}.

\section{Quantization-Aware Training} \label{sec:qat}
Post-training quantization (PTQ)~\cite{jacob2018quantization} applies quantization after a model is fully trained. 
In contrast, quantization-aware training (QAT)~\cite{nagel2022overcoming} incorporates quantization effects during training, allowing the model to adapt its parameters to the quantization error and thereby improving low-bit accuracy at inference time.

In \our, we further accelerate the sparse attention branch $O_s$ computation using a low-bit attention in a QAT manner. Concretely, during training, we use low-bit attention \emph{only in the forward pass}, while the backward pass remains fully in FP16. This design enables the attention speedup brought by low-bit attention while minimizing the end-to-end accuracy drop caused by low-bit quantization.

\paragraph{Forward (low-bit attention).}
Given $Q,K,V\in\mathbb{R}^{N\times d}$, we apply a low-bit quantized attention in the forward pass. We first quantize $Q$ ($\hat Q, s_Q = \mathrm{quant}(Q)$) and $K$ ($\hat K, s_K = \mathrm{quant}(K)$), then compute
\begin{align}
S=\mathrm{dequant}({\hat Q\hat K^\top}/{\sqrt{d}},\, s_Q,\, s_K), \notag ~~~P=\mathrm{softmax}_{M}(S),
\end{align}
followed by quantizing $P$ ($\hat P, s_P = \mathrm{quant}(P)$) and $V$ ($\hat V, s_V = \mathrm{quant}(V)$) and computing
\begin{equation} \notag
O_s=\mathrm{dequant}(\hat P\hat V,\, s_P,\, s_V).
\end{equation}
Here, $\mathrm{quant}(\cdot)$ maps an FP16 tensor to a low-bit tensor (e.g., INT8 or FP8) along with its scale, and $\mathrm{dequant}(\cdot)$ rescales the result back to FP16. We use $\hat Q,\hat K,\hat P,\hat V$ to denote the quantized tensors and $s_Q,s_K,s_P,s_V$ to denote their scales. Our quantization/dequantization scheme follows SageAttention2++~\cite{zhang2025sageattention2++}.

Note that the equations above describe the mathematical computation rather than the GPU kernel implementation. We build the actual efficient kernel on the FlashAttention algorithm to avoid computing the full score matrix $S$ before applying mask $M$. Instead, we skip unnecessary computations. The detailed algorithm is provided in Sections~\ref{sec:training} and~\ref{sec:inference}.

\paragraph{Backward (FP16-only).}
Let $dO_s$ denote the gradient of $O_s$. In our QAT setting, the backward pass is computed entirely in FP16, using the original FP16 inputs $(Q,K,V)$ and the forward output $O_s$. The gradient of $Q, K, V$ from the sparse attention branch can be formulated as:
\begin{equation} \notag
dQ,\ dK,\ dV\ =\ \mathrm{backward}(dO_s,\ O_s,\ Q,\ K,\ V).
\end{equation}

The detailed backward GPU kernel, along with the complete training pipeline, is provided in Section~\ref{sec:training}.

\begin{algorithm}[h!]
    \small
    \caption{Forward pass of \our.}
    \label{alg:fwd} 
    \begin{algorithmic}[1]
    \STATE {\bf Input:} {Matrices $Q, K, V \in \mathbb{R}^{N \times d}$, $b_q, b_{k}$, k\%, learnable \add{$\mathrm{proj}_q$, $\mathrm{proj}_k \in \mathbb{R}^{d \times d}$, $\alpha \in \mathbb{R}^{N/b_q \times 1}$}}.
    
    \STATE $K = K - \rm{colmean}(K)$ ;  ~//~ \annotate{smooth K of SageAttention}
    
    \STATE $ Q^\phi, K^\phi = \phi(Q), \phi(K) , ~~~ \bar Q, \bar K = \rm{pool}(Q), \rm{pool}(K) ; $
    \STATE {Divide $Q, Q^\phi$ to $T_m = \frac{N}  {b_{q}} $ blocks $\{\vQ_i\}$, $\{\vQ^\phi_i\}$} ; ~{Divide $K, V, K^\phi$ to $T_n\text{=} \frac{N}  {b_{k}}$ blocks $\{\vK_i\}$, $\{\vV_i\}$, $\{\vK_i^\phi\}$};
    \STATE $h=\{h_j\}=\{(\vK_j^\phi)^\top \vV_j\}$ ; 
    ~ $z=\{z_j\}=\{{\rm rowsum}((\vK_j^\phi)^\top)\}$ ; ~$M_c[:,:]=0$ ; 

    \STATE {$P_c = {\rm softmax}(\add{\rm{proj_q}} (\bar Q) \add{\rm{proj}_k}(\bar K)^\top / \sqrt{d})$ ;} 
    ~ $M_c = \rm{Top}\text{-}\rm{k}(k\%, P_c)$ ; ~//~ \annotate{\add{SoftTop-k} in stage1 training}

    \FOR {$i=1$ {\bf to} $T_m$}
        \STATE Initialize $m_{i,0}=-\infty$, $l_{i,0}=0$, $\vO^s_{i,0}=0$, $\vH_i=0$, $\vZ_i=0$.
        \FOR {$j=1$ {\bf to} $T_n$} 
            \IF {$M_c[i,j]=1$}
                \STATE {$\vS_{ij} = \add{\rm{dequant}}(\add{\rm{quant}}(\vQ_i) \add{\rm{quant}}(\vK_j)^\top) / \sqrt{d}$; 
                ~ $m_{ij} = {\rm max}(m_{i, j-1}, {\rm rowmax}(\vS_{ij}))$} ;
                
                \STATE {$\vP_{ij}=\exp(\vS_{ij}-m_{ij})$} ;
                ~ {$l_{ij}=e^{m_{i,j-1}-m_{ij}} l_{i,j-1} + {\rm rowsum}(\vP_{ij})$ ; 

                \STATE $O_{\rm{tmp}} = \add{\rm{dequant}}(\add{\rm{quant}}(\vP_{ij}) \add{\rm{quant}}(\vV_j))$ ;
                ~ $\vO_{ij}^s = {\rm diag}(e^{m_{i,j-1}-m_{ij}}) \vO_{i,j-1}^s + O_{\rm{tmp}}$} ;
            \ELSIF {$M_c[i,j]=0$}
                \STATE Carry $m_{ij},l_{ij},\vO^s_{ij}$ from tile $j-1$.
                \STATE $\vH_i \gets \vH_i + h_j ; ~~~~~ \vZ_i \gets \vZ_i + z_j$ ;
            \ENDIF
        \ENDFOR
        \STATE $\vO_i^s={\rm diag}(l_{i,T_n})^{-1}\vO_{i,T_n}^s$ ; 
        ~ $\vO_i^l = \vQ_i^\phi \vH_i / (\vQ_i^\phi \vZ_i) ; ~~~ \mathbf{L}_i = m_{i, T_n} + \mathrm{log}(l_{i, T_n})$ ;
    \ENDFOR

    \STATE $O^s=\{\vO^s_i\}$,~~~$O^l=\{\vO^l_i\}$ ;
    \STATE \textbf{return} $O \ =\ \add{\alpha}\odot O^s\ +\ (1-\add{\alpha})\odot O^l$ ;
    \end{algorithmic}
\end{algorithm}

\section{Training with \our} \label{sec:training}

To fine-tune a diffusion model with \our, we adopt a two-stage training strategy. \scircled{1} In the first stage, we seek a better initialization for $\mathcal{R}$ and $\alpha$ to ensure stable and effective subsequent fine-tuning of the diffusion model. \scircled{2} In the second stage, we fine-tune the entire diffusion model end-to-end. In this stage, we directly optimize the diffusion loss over all model parameters $\Theta$, including $\alpha$, without $\mathcal{R}$, so that the model adapts to high-sparsity attention and can even achieve better performance under high sparsity.

Specifically, in the first stage, we use the $Q$, $K$, and $V$ matrices from every attention layer at each diffusion timestep as training data. For each sparsity setting (i.e., different $k\%$, we use 5\%, 4\%, and 3\%), we train $\mathcal{R}$ and $\alpha$. Note that Top-k is non-differentiable. Therefore, during Stage~1, we replace the Top-k operator in Equation~\ref{equ:mask_of_sla2} with a SoftTop-k operator~\citep{ding2024separatedynamicdifferentiablesmart}:
\begin{equation}
\mathrm{SoftTop}\text{-}{\rm k}(k\%, P_c)_{ij} = \sigma\left(\frac{(P_c)_{ij}}{\tau} + \lambda_i\right),
\end{equation}
where $\sigma$ denotes the sigmoid function, $\tau$ is a temperature parameter, and $\lambda_i$ is solved via binary search to ensure that each row sums to $k\% \times N/b_k$. The gradient of SoftTop-k is computed using the reparameterization trick (see~\cite{ding2024separatedynamicdifferentiablesmart}), which enables gradient backpropagation. This operator retains key properties of Top-k, such as enforcing a row-wise sum of $k\% \times N/b_k$.
The overall training algorithm is in Algorithms~\ref{alg:training}, where we use $O = \our(Q, K, V, k\%, \mathcal{R}, \alpha)$ as \our operator. The forward and backward procedures of \our, are provided in Algorithms~\ref{alg:fwd}, and~\ref{alg:bwd}, respectively.

\section{Inference with \our}  \label{sec:inference}

During inference, we simply replace the attention modules in the diffusion model with \our and run the \our forward pass described in Algorithm~\ref{alg:fwd}. Note that the Top-$k$ operation uses the hard Top-$k$ in Equation~\ref{equ:mask_of_sla2}, rather than SoftTop-$k$.

\begin{table*}[ht]
    \centering
    \caption{Quality and efficiency metrics of \our and the baseline methods. Bold quality scores indicate the best result within each sparse-method group.}
    \small
    \label{table:exp_effectiveness}
    \setlength\tabcolsep{5pt}
    \begin{tabular}{cccccccccc}
    \toprule
    \multirow{2}{*}{\textbf{Model}} &
    \multirow{2}{*}{\textbf{Method}} &
    \multicolumn{6}{c}{\textbf{Quality}} & \multicolumn{2}{c}{\textbf{Efficiency}} \\
    \cmidrule(lr){3-8} \cmidrule(lr){9-10} && \texttt{IQ} $\uparrow$ & \texttt{OC} $\uparrow$ & \texttt{AQ} $\uparrow$ & \texttt{MS} $\uparrow$ & \texttt{SC} $\uparrow$ & \texttt{VR} $\uparrow$ & \texttt{FLOPs} $\downarrow$ & Sparsity $\uparrow$ \\
    \midrule
    \multirow{10}{*}{\parbox{1cm}{\centering \textbf{Wan2.1\\-T2V\\-1.3B\\-480P}}} &
    Full Attention & 63.67 & 20.27 & 64.41 & 98.95 & 95.40 & 0.1084  & 52.75T & 0\% \\
    \cmidrule(lr){2-10}
    & \texttt{VMoBA} & 65.31 & 20.82 & 64.14 & 97.80 & 86.69 & 0.0936  & 5.28T & \multirow{4}{*}{90\%} \\
    & \texttt{VSA}   & 59.57 & 19.27 & 50.60 & 97.44 & 87.98 & -0.0881 & 5.40T \\
    & \texttt{SLA}   & 63.10 & 20.88 & 64.34 & 97.90 & 92.54 & 0.0872  & 5.40T \\
    & \our & \cellcolor{lightblue} \textbf{67.70} & \cellcolor{lightblue} \textbf{21.62} & \cellcolor{lightblue} \textbf{64.86} & \cellcolor{lightblue} \textbf{98.69} & \cellcolor{lightblue} \textbf{95.54} & \cellcolor{lightblue} \textbf{0.1093} & 5.51T \\
    \cmidrule(lr){2-10}
    & \texttt{VMoBA} & 63.08 & 21.07 & 61.96 & 97.68 & 79.83 & 0.0746  & 2.64T & \multirow{4}{*}{95\%} \\
    & \texttt{VSA}   & 55.50 & 14.95 & 42.13 & 96.19 & 88.34 & -0.1309 & 2.75T \\
    & \texttt{SLA}   & 63.14 & 21.09 & 62.91 & 97.83 & 94.36 & 0.0881  & 2.75T \\
    & \our & \cellcolor{lightblue} \textbf{67.04} & \cellcolor{lightblue} \textbf{21.55} & \cellcolor{lightblue} \textbf{64.90} & \cellcolor{lightblue} \textbf{98.46} & \cellcolor{lightblue} \textbf{95.27} & \cellcolor{lightblue} \textbf{0.1023} & 2.87T \\
    \cmidrule(lr){2-10}
    & \our & \cellcolor{lightblue} \textbf{66.64} & \cellcolor{lightblue} \textbf{21.42} & \cellcolor{lightblue} \textbf{64.62} & \cellcolor{lightblue} \textbf{98.04} & \cellcolor{lightblue} \textbf{94.83} & \cellcolor{lightblue} \textbf{0.1039}  & \cellcolor{lightblue} \textbf{1.82T} & \cellcolor{lightblue} \textbf{97}\% \\
    \midrule
    
    \multirow{10}{*}{\parbox{1cm}{\centering \textbf{Wan2.1\\-T2V\\-14B\\-720P}}} &
    Full Attention & 68.01 & 22.44 & 64.66 & 99.14 & 95.93 & 0.1238  & 292.6T & 0\% \\
    \cmidrule(lr){2-10}
    & \texttt{VMoBA} & 67.18 & 20.85 & 63.64 & 98.55 & 94.50 & 0.1117  & 29.26T & \multirow{4}{*}{90\%} \\
    & \texttt{VSA}   & 64.03 & 21.27 & 63.37 & \textbf{98.90} & 93.65 & 0.1074 & 20.92T \\
    & \texttt{SLA}   & 67.58 & \textbf{21.62} & 63.80 & 98.78 & \textbf{95.74} & 0.1166  & 20.92T \\
    & \our & \cellcolor{lightblue} \textbf{69.63} & \cellcolor{lightblue} 20.68 & \cellcolor{lightblue} \textbf{66.41} & \cellcolor{lightblue} 98.84 & \cellcolor{lightblue} \textbf{95.74} & \cellcolor{lightblue} \textbf{0.1238} & 21.16T \\
    \cmidrule(lr){2-10}
    & \texttt{VMoBA} & 21.27 & 7.96 & 33.59 & \textbf{99.99} & \textbf{100} & -0.0965  & 14.63T & \multirow{4}{*}{95\%} \\
    & \texttt{VSA}   & 47.69 & 13.90 & 34.95 & 97.09 & 91.12 & -0.1822 & 14.87T \\
    & \texttt{SLA}   & 64.43 & 20.89 & 61.89 & 98.86 & 94.41 & 0.1078  & 14.87T \\
    & \our & \cellcolor{lightblue} \textbf{69.02} & \cellcolor{lightblue} \textbf{21.11} & \cellcolor{lightblue} \textbf{65.55} & \cellcolor{lightblue} 98.89 & \cellcolor{lightblue} 95.53 & \cellcolor{lightblue} \textbf{0.1125} & 15.11T \\
    \cmidrule(lr){2-10}
    & \our & \cellcolor{lightblue} \textbf{66.93} & \cellcolor{lightblue} \textbf{21.12} & \cellcolor{lightblue} \textbf{65.14} & \cellcolor{lightblue} \textbf{98.71} & \cellcolor{lightblue} \textbf{94.42} & \cellcolor{lightblue} \textbf{0.1149}  & \cellcolor{lightblue} \textbf{9.26T} & \cellcolor{lightblue} \textbf{97\%} \\
    \bottomrule
    \end{tabular}
\end{table*}

\begin{figure*}[h]
    \centering
    \includegraphics[width=0.999\linewidth]{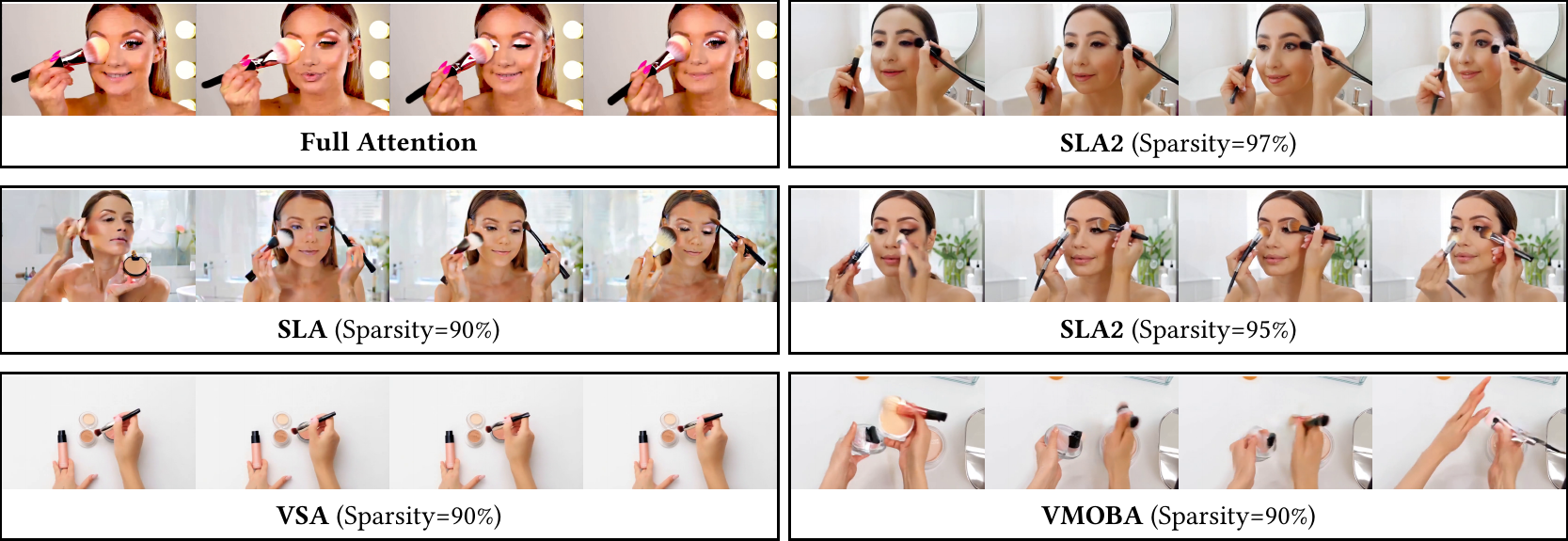}
    \vspace{.5em}
    \caption{Visible examples of \our and baselines on Wan2.1-T2V-1.3B-480P model. The prompt used for generation is in Appendix~\ref{sec:prompts}.}
    \label{fig:visible_examples}
\end{figure*}

\section{Insights} \label{sec:insight}
We summarize key insights on SLA design and training in a question-driven format.

\vspace{1em}
\textbf{(1) Why is the design of $\mathcal{R}$ (Equation~\ref{equ:mask_of_sla2}) reasonable?}
The core motivation of sparse-linear attention is to decompose the attention weights as $P=P_1+P_2$, where $P_1$ is handled by the sparse branch, and $P_2$ is handled by the linear branch. It aims to route a low-rank portion of $P$ to $P_2$ and make $P_1$ as sparse as possible without harming end-to-end quality. We explain the design choices of $\mathcal{R}$ by answering three sub-questions:

\underline{\textbf{(1.a)} Why the input of $\mathcal{R}$ are $Q$ and $K$?}
For each attention layer, the attention weights are determined by the score matrix $S=QK^\top/\sqrt{d}$ followed by a row-wise softmax, i.e., $P=\mathrm{Softmax}(S)$. Therefore, deciding which positions of $P$ should be assigned to the sparse branch is fundamentally a decision about which positions of $S$, i.e., the matrix multiplication between $Q$ and $K$, are likely to contribute most after softmax. This makes $(Q,K)$ the natural and sufficient signals for routing, while $V$ does not affect the formation of $P$ and is thus not needed for the routing decision.

\underline{\textbf{(1.b)} Why apply pooling to $Q$ and $K$ in $\mathcal{R}$?}
A naive router that operates on the full $QK^\top$ would incur $\mathcal{O}(N^2)$ complexity, which is too expensive. To reduce this cost, we pool adjacent tokens in $Q$ and $K$ using mean pooling to obtain $\bar Q$ and $\bar K$. This is still effective because nearby tokens in diffusion transformers often have similar distribution~\cite{jiang2024minference,zhangspargeattention,gao2024seerattention}, so the values in $QK^\top$ vary smoothly across adjacent positions. 

\underline{\textbf{(1.c)} Why using projections ($\rm{proj}_q$ and $\rm{proj}_k$) in $\mathcal{R}$?}
Using $\bar Q\bar K^\top$ followed by softmax and Top-$k$ is a simple heuristic and may not yield an optimal split of $P$ into a sparse part and a low-rank part. By introducing learnable projections $\mathrm{proj}_q$ and $\mathrm{proj}_k$, the router can learn a task-adaptive representation in which Top-$k$ selection better matches the desired decomposition (making $P_1$ much sparser while leaving a portion that is easier for the linear branch to approximate). In particular, this design generalizes the heuristic routing: setting $\mathrm{proj}_q=\mathrm{proj}_k=I$ recovers the original form, while learning these projections under our training objective can produce a more suitable partition.

\vspace{1em}
\textbf{(2) Why does \our need two-stage training?}
We adopt a two-stage training strategy for both training stability and train--inference consistency. First, before end-to-end fine-tuning of the entire diffusion model, $\mathcal{R}$ should be reasonably initialized. Otherwise, unstable and poor routing can make subsequent fine-tuning difficult. Second, the router used at inference relies on hard Top-k, which is non-differentiable and blocks gradient propagation. To train the projection parameters inside $\mathcal{R}$, we therefore use a differentiable SoftTop-$k$ operator during Stage~1. After obtaining a good initialization, Stage~2 fine-tunes the full diffusion model while keeping the routing computation aligned with inference (i.e., using hard Top-k for routing), ensuring that the trained model matches the inference-time computation logic.

\begin{figure*}[h]
    \centering
    \includegraphics[width=0.995\linewidth]{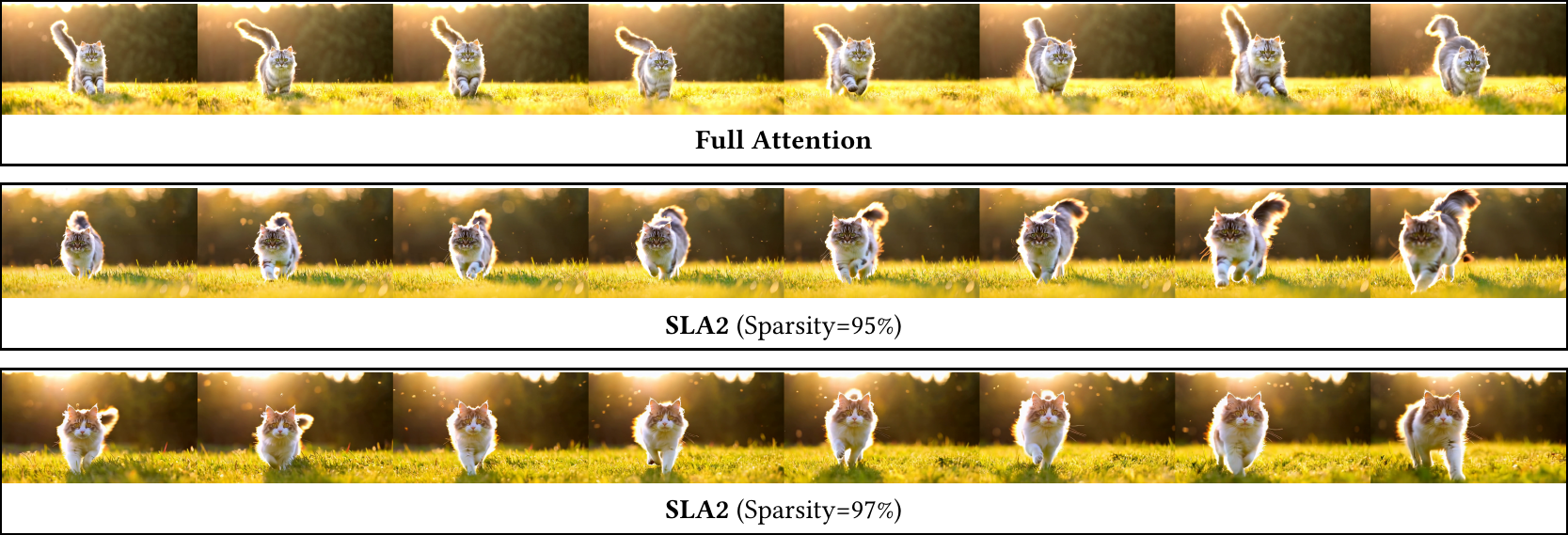}
    \caption{Visible examples of \our and baselines on Wan2.1-T2V-14B-720P model. The prompt used for generation is in Appendix~\ref{sec:prompts}.}
    \vspace{-1.5em}
    \label{fig:visible_examples_14B}
\end{figure*}

\section{Experiments}
\vspace{-.5em}
\subsection{Setup} \label{sec:exp_setup} 
\textbf{Model and Baselines.} We fine-tune \our and baseline methods on the Wan2.1-1.3B-480P and Wan-2.1-14B-720P models~\citep{wan2025}. For the dataset, we use a private video dataset of 3,000 videos (about 5 seconds each) collected from public sources. To construct text--video pairs, we generate a caption for each video using Qwen3-VL-Flash and use these captions as text conditioning for both fine-tuning and evaluation. For baselines, we use Full Attention (without training) implemented with FlashAttn2. We also select several state-of-the-art video generation methods with sparse attention mechanism, including \texttt{SLA}~\citep{zhang2025sla}, \texttt{VSA}~\citep{zhang2025vsa} and \texttt{VMoBa}~\citep{wu2025vmoba}. All results are obtained using the official open-source implementations.

\textbf{Hyper-parameters.} We fine-tune each method for 500 steps. The batch size is set to 64 for the 1.3B model and 15 for the 14B model. We set the block sizes to $b_q = 128$ and $b_{kv} = 64$. We use $k\%$ of 5\%, 4\%, and 3\% for \our. For the temperature parameter $\tau$ in SoftTop-k, we use $\tau=0.1$.

\textbf{Metrics.} Following \cite{zhang2024evaluationagent,yang2024cogvideox}, we evaluate video quality using multiple dimensions from VBench~\citep{huang2024vbench}, including Imaging Quality ({\texttt{IQ}}), Overall Consistency ({\texttt{OC}}), Aesthetic Quality ({\texttt{AQ}}), Motion Smoothness ({\texttt{MS}}) and Subject Consistency ({\texttt{SC}}). In addition, we assess human preference using the Vision Reward metric ({\texttt{VR}})~\citep{xu2024visionreward}. To quantify computational cost, we use {\texttt{FLOPs}} (floating-point operations). For kernel-level efficiency, we report $C / t$, where $C = 4N^2 d$ denotes the theoretical amount of computation and $t$ is the execution latency. We measure the end-to-end inference latency in seconds.

\subsection{Effectiveness} \label{sec:video_effectiveness}
Table~\ref{table:exp_effectiveness} compares the video generation quality and efficiency of \our against baseline methods on the Wan2.1-T2V-1.3B-480P and Wan2.1-T2V-14B-720P models. At sparsity levels of 90\% and 95\%, \our improves IQ and AQ over all sparse baselines on both models. At 97\% sparsity, \our retains higher AQ than all sparse baselines at 90\% sparsity on both models, while reducing attention FLOPs by approximately $\mathbf{29}\times$--$\mathbf{32}\times$ relative to Full Attention. Interestingly, we observe that sparse attention methods can even outperform Full Attention on many metrics after fine-tuning. The matched fine-tuning comparison in Appendix~\ref{sec:matched-finetuning} controls for the benefit of fine-tuning.

\textbf{Visible examples.}
Figure~\ref{fig:visible_examples} shows an example generated by different methods fine-tuned on Wan2.1-T2V-1.3B-480P. The videos produced by \our exhibit the highest quality and maintain content similar to that generated by Full Attention. In contrast, videos from other methods either differ noticeably from Full Attention or show clear distortions. Figure~\ref{fig:visible_examples_14B} presents an example generated by Full Attention and \our on Wan2.1-T2V-14B-720P model. \our brings almost no degradation in video quality.

\begin{figure}[h!]
    \centering

    \begin{minipage}[t]{0.42\columnwidth}
        \centering
        \vspace{0pt}
        \includegraphics[width=\linewidth]{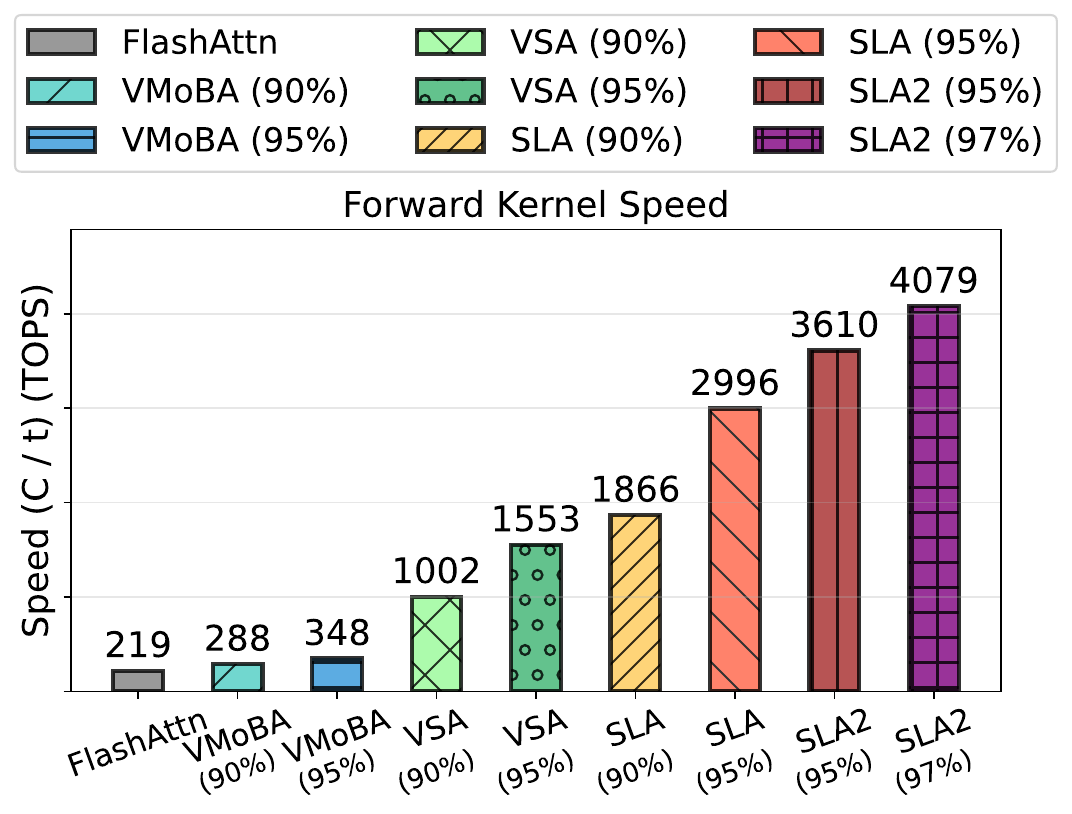}
        \caption{Kernel speed of \our and baselines with different sparsities.}
        \label{fig:speed_comparison}
    \end{minipage}
    \hfill
    \begin{minipage}[t]{0.56\columnwidth}
        \centering
        \vspace{0pt}
        \captionof{table}{Ablation experiments results.}
        \label{table:exp_ablation}
        \small
        \setlength\tabcolsep{3pt}
        \begin{tabular}{ccccccc}
        \toprule
        \multirow{2}{*}{\textbf{Method}} 
        & \multicolumn{6}{c}{\textbf{Quality}} \\
        \cmidrule(lr){2-7} 
        & \texttt{IQ} $\uparrow$ 
        & \texttt{OC} $\uparrow$ 
        & \texttt{AQ} $\uparrow$ 
        & \texttt{MS} $\uparrow$ 
        & \texttt{SC} $\uparrow$ 
        & \texttt{VR} $\uparrow$ \\
        \midrule
        Full Attention & 63.67 & 20.27 & 64.41 & 98.95 & 95.40 & 0.1084 \\
        \midrule
        w/o QAT     & 65.28 & 20.66 & 61.85 & 97.44 & 94.65 & 0.0850 \\
        Topk-router & 63.66 & 20.9  & 62.65 & 97.86 & 94.26 & 0.0876 \\
        \rowcolor{lightblue}
        \our        & \textbf{66.64} & \textbf{21.42} & \textbf{64.62} & \textbf{98.04} & \textbf{94.83} & \textbf{0.1039} \\
        \midrule
        \our (85\%) & 67.97 & 21.98 & 64.79 & 98.75 & 95.79 & 0.1135 \\
        \our (90\%) & 67.70 & 21.62 & 64.86 & 98.69 & 95.54 & 0.1093 \\
        \our (95\%) & 67.04 & 21.55 & 64.9  & 98.46 & 95.27 & 0.1023 \\
        \our (97\%) & 66.64 & 21.42 & 64.62 & 98.04 & 94.83 & 0.1039 \\
        \bottomrule
        \end{tabular}
    \end{minipage}

    \vspace{-.5em}
\end{figure}

\begin{figure}[h!]
    \centering
    \includegraphics[width=0.999\columnwidth]{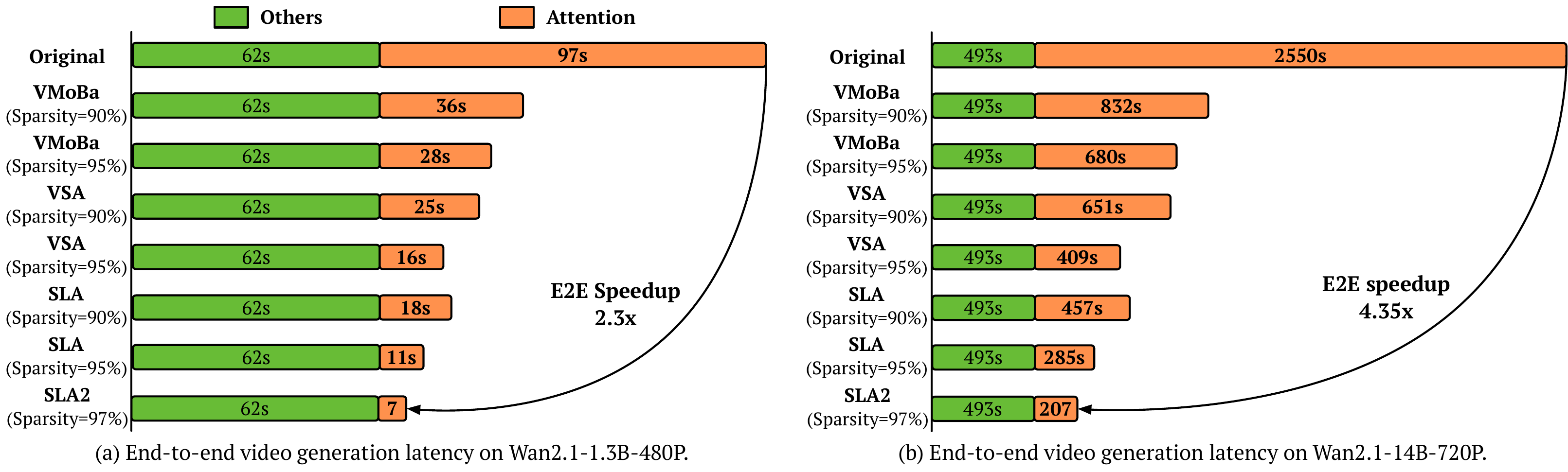}
    \caption{End-to-end generation latency of \our and baselines with different sparsities.}
    \vspace{-1.5em}
    \label{fig:e2e_latency}
\end{figure}

\subsection{Efficiency}
Figure~\ref{fig:speed_comparison} illustrates the forward kernel speed of \our and the baseline methods on an RTX5090, measured in TOPS (trillion operations per second). At 97\% sparsity, \our achieves a $\mathbf{18.6}\times$ speedup over FlashAttn2, and is $11.7\times$ and $2.6\times$ faster than VMoBA and VSA at 95\% sparsity, respectively. This comparison combines the effects of higher sparsity and the low-bit sparse-attention kernel. 
Figure~\ref{fig:e2e_latency} presents the end-to-end video generation latencies for \our and the baselines. On the Wan-1.3B-480P model, reducing attention latency from 97s to 7s ($\mathbf{13.9}\times$ speedup) enables \our to achieve a $\mathbf{2.30}\times$ reduction in overall end-to-end latency. On the Wan-14B-720P model, \our further reduces end-to-end latency by $\mathbf{4.35}\times$. Since the Wan2.1-14B-720P model exceeds the VRAM capacity of a single RTX5090, we enable sequential CPU offloading during evaluation. The reported latency already excludes the offloading overhead.

\subsection{Ablation Study}
\label{sec:ablation}

\textbf{Quantization-aware training.} To evaluate the impact of quantization-aware training (QAT), we fine-tune the same model without QAT and perform quantized inference. As shown in Table~\ref{table:exp_ablation}, the quality of generated videos drops when quantized inference follows training without QAT, which confirms its effectiveness. For efficiency, we evaluate \our both with and without quantization. Low-bit quantization provides an approximately 1.3x kernel speedup.

\textbf{Learnable router.} 
To evaluate the benefit of the learnable router, we compare it with the Top-k router used in SLA~\citep{zhang2025sla}, which directly selects the largest scores in ${\rm pool}(Q){\rm pool}(K)^\top$. As shown in Table~\ref{table:exp_ablation}, the learnable router significantly outperforms the Top-k router.

\textbf{Varying sparsity.} 
We vary the sparsity from $85\%$ to $97\%$ and evaluate \our under different sparsity levels. As summarized in Table~\ref{table:exp_ablation}, lower sparsity generally improves quality, although AQ and VR are not monotonic. At $97\%$ sparsity, \our remains competitive with the sparse baselines in Table~\ref{table:exp_effectiveness}.

\textbf{Mixing coefficient.}
Removing the $\alpha$-based correction at 97\% sparsity on Wan2.1-1.3B degrades all six reported quality metrics (Appendix~\ref{sec:alpha-ablation}), supporting the contribution of the mixing formulation.

\textbf{Matched fine-tuning.}
When both models are fine-tuned on the same 3,000-video dataset for 500 steps, SLA2 at 90\% sparsity outperforms full attention on all three reported metrics (Appendix~\ref{sec:matched-finetuning}).

\section{Related Work}

Sparse attention and linear attention are two main ways to speed up attention in Transformer-based models. Sparse attention methods can be grouped by whether they require training. Training-free approaches~\citep{xiao2023efficient,jiang2024minference,gao2024seerattention,xi2025sparse,zhangspargeattention,ribar2023sparq,yang2025sparse,li2025radial,chen2025sparse,lai2025flexprefill,zhang2023h2o,tang2024quest,zhu2025tactic,lin2025twilight,xu2025xattention,xia2025training,chen2025re,zhang2025fast,yang2024post} reduce inference cost by masking attention patterns at test time, while trainable methods~\citep{zhang2025vsa,wu2025vmoba,zhang2025sla,zhan2025bidirectional,zhou2025trainable,lu2025moba,yuan2025native,liu2025deepseek,cai2025mixture,liu2025fpsattention,sun2025vorta,tan2025dsv,ding2023longnet} encourage sparsity during training and can support higher sparsity. Linear attention methods~\citep{wang2020linformer,choromanski2020rethinking,katharopoulos2020transformers,qin2024lightning,yang2024gated,sun2023retentive} are mainly studied in language models. In diffusion transformers, SANA~\citep{xie2024sana} and Dig~\citep{zhu2025dig} show that linear attention can work for image-generation pre-training; however, for video generation, linear attention alone often cannot keep quality. In addition, hardware-focused work~\citep{dao2022flashattention,dao2023flashattention,shah2024flashattention,zhang2025sageattention,zhang2025sageattention2,zhang2025sageattention3} speeds up attention by improving GPU execution through tiling, kernel fusion, and quantization.

SpargeAttention2~\citep{ConcurrentWork1} combines hybrid Top-k+Top-p masking with distillation fine-tuning; SVG-EAR~\citep{zhou2026svg} uses error-aware routing and parameter-free linear compensation.

Low-bit attention~\citep{zhang2025sageattention,zhang2025sageattention2,zhang2025sageattention2++,zhang2025sageattention3,zhang2026sagebwd}, accurate INT8 training~\citep{zhang2025accurate}, and weight-sensitivity analysis~\citep{hu2025identifying} provide complementary efficiency techniques. Quant VideoGen~\citep{xi2026quant} targets KV-cache quantization, while 6Bit-Diffusion~\citep{su2026bit} studies mixed-precision video diffusion. Broader acceleration directions include diffusion distillation~\citep{zheng2025large,zheng2026causal,yang2026turbot2va}, model stitching~\citep{sun2026hybridstitch}, speculative decoding~\citep{hu2026speculative}, and TurboDiffusion~\citep{zhang2025turbodiffusion}. Resolution extrapolation~\citep{zhao2025ultravico,zhao2025ultraimage}, interactive video generation~\citep{zhang2026vidu,zhang2026vidus2}, and streaming serving~\citep{jiang2026turboserve} further motivate efficient attention at scale. A broader taxonomy is given by the efficient-attention survey~\citep{zhangefficient}.

\section{Conclusion}
We presented \our, a trainable sparse-linear attention method for diffusion models. It is motivated by two limitations of SLA: its heuristic routing based on the magnitude of attention weights and a mismatch with the decomposition of sparse and linear attention, revealed by our error analysis.
\our addresses these issues by introducing a learnable router and a decomposition-consistent mixing formulation. Moreover, \our adopts a sparse + low-bit attention in a quantization-aware fine-tuning way for further acceleration.
Experiments show that \our achieves up to $97\%$ attention sparsity and an $18.6\times$ attention speedup, while preserving video generation quality. We hope \our offers an effective and practical way for efficient attention in diffusion models.

\begin{ack}
This work was supported by the National Natural Science Foundation of China (Nos. 62550004, 62595773, and 62376131) and the Ministry of Education of China (No. JYB2025XDXM101).
The authors declare no other competing interests.
\end{ack}

\bibliographystyle{unsrt}
\bibliography{main,camera-ready-references}

\clearpage
\appendix
\section{Additional Experiments}
\label{app:additional-experiments}

\subsection{Mixing Coefficient}
\label{sec:alpha-ablation}
Following the Wan2.1-1.3B-480p ablation setup at 97\% sparsity in Table~\ref{table:exp_ablation}, we remove the $\alpha$-based correction while keeping the remaining setup unchanged. Table~\ref{tab:alpha-ablation} shows that the correction improves all six reported quality metrics, supporting the contribution of the mixing formulation beyond the other components.

\begin{table}[htbp]
\centering
\caption{Ablation of the $\alpha$-based correction. Higher scores are better.}
\label{tab:alpha-ablation}
\begin{tabular}{lrrrrrr}
\toprule
Setting & IQ & OC & AQ & MS & SC & VR \\
\midrule
Without $\alpha$ & 64.05 & 20.66 & 61.79 & 94.71 & 91.37 & 0.1008 \\
With $\alpha$ & \textbf{66.64} & \textbf{21.42} & \textbf{64.62} & \textbf{98.04} & \textbf{94.83} & \textbf{0.1039} \\
\bottomrule
\end{tabular}
\end{table}

\subsection{Comparison under Matched Fine-Tuning}
\label{sec:matched-finetuning}
We fine-tune the full-attention model and SLA2 for 500 steps on the same 3,000-video dataset, using Wan2.1-1.3B at 480p and 90\% sparsity for SLA2. As shown in Table~\ref{tab:matched-finetuning}, SLA2 outperforms the fine-tuned full-attention baseline on all three reported metrics.

\begin{table}[htbp]
\centering
\caption{Comparison under matched fine-tuning data and step budgets. Higher scores are better.}
\label{tab:matched-finetuning}
\begin{tabular}{lrrrr}
\toprule
Method & Fine-tuning steps & IQ & OC & VR \\
\midrule
Full Attention & 0 & 63.67 & 20.27 & 0.1084 \\
Full Attention & 500 & 65.79 & 21.10 & 0.1014 \\
SLA2 (90\% sparsity) & 500 & \textbf{67.70} & \textbf{21.62} & \textbf{0.1093} \\
\bottomrule
\end{tabular}
\end{table}

\section{Backward Pass of \our}

The backward pass of \our is presented in Algorithm~\ref{alg:bwd}. Following SLA~\citep{zhang2025sla}, we manually derive the gradients with respect to $Q, K, V, Q^\phi$ and $K^\phi$, while all remaining gradients are computed via PyTorch’s automatic differentiation. Note that ${\vdH_i}$ and ${\vdZ_i}$ are precomputed, such that the main procedure involves only a single matrix addition (Line 15), thereby improving computational efficiency.

\begin{algorithm}[H]
    \small
    \caption{Backward pass of \our.}
    \label{alg:bwd} 
    \begin{algorithmic}[1]
    \STATE {\bf Input:} {$Q, K, V, Q^\phi, K^\phi, M_c, \{\mathbf L_i\}, \{\vH_i\}, \{\vZ_i\}, O^s, O^l $ from the forward, $dO^s,dO^l \in \mathbb R^{N\times d}$}.
    \STATE Initialize $dQ,dK,dV$ to zero.
    \STATE {$D^s={\rm rowsum}(dO^s\odot O^s)$, $D^l={\rm rowsum}(dO^l\odot O^l)$, divide $D^s, D^l$ into $T_m$ blocks $\{\vD_i^s\},\{\vD_i^l\}$} ;
    \FOR {$i=1$ {\bf to} $T_m$}
        \STATE {$\vdH_i=(\vQ_i^\phi/(\vQ^\phi_i \vZ_i))^\top \vdO^l_i$; ~ $\vdZ_i=-(\vQ_i^\phi/(\vQ_i^\phi \vZ_i))^\top D_i^l$} ;
        \STATE {$\vdQ^\phi_i=(\vdO^l_i (\vH_i)^\top - \vD_i^l\vZ_i^\top) / (\vQ^\phi_i \vZ_i)$} ;
    \ENDFOR
    \FOR {$j=1$ {\bf to} $T_n$}
        \STATE {Initialize $\vdH=0,\vdZ=0$} ;
        \FOR {$i=1$ {\bf to} $T_m$}
            \IF {$M_c[i,j]=1$}
                \STATE {$\vS_{ij} = \vQ_i \vK_j^\top / \sqrt{d}$} ;~~ {$\vP_{ij}=\exp(\vS_{ij}-\mathbf L_i)$ ; ~~ $\vdV_j\gets\vdV_j + \vP_{ij}^\top\vdO_i^s$} ;~~ $\vdP_{ij}=\vdO^s_i\vV_j^\top$ ;
                \STATE {$\vdS_{ij}=\vP_{ij}\odot(\vdP_{ij}-\vD_i^s)$} ;~~~ {$\vdQ_i\gets\vdQ_i + \vdS_{ij}\vK_j/\sqrt{d}$ ; ~~~~~ $\vdK_j\gets\vdK_j + \vdS_{ij}^\top\vQ_i/\sqrt{d}$} ;
            \ELSIF {$M_c[i,j]=0$}
                \STATE $\vdH \gets \vdH + \vdH_i ; ~~~~~ \vdZ \gets \vdZ + \vdZ_i$ ;
            \ENDIF
        \ENDFOR
        \STATE {$\vdK^\phi_j=\vV_j(\vdH)^\top+(\vdZ)^\top ; ~~~~~ \vdV_j\gets\vdV_j+\vK^\phi_j\vdH$} ;
    \ENDFOR
    \STATE \textbf{return} $dQ=\{\vdQ_i\}$,~~~$dK=\{\vdK_i\}$,~~~$dV=\{\vdV_i\}$,~~~$dQ^\phi=\{\vdQ^\phi_i\}$,~~~$dK^\phi=\{\vdK^\phi_i\}$ ;
    \end{algorithmic}
\end{algorithm}

\section{Prompts Used} \label{sec:prompts}

The prompt we used for Figure~\ref{fig:visible_examples} is: \textit{``A first-person perspective video of a morning makeup routine in a bright, minimalist bathroom. The hands apply moisturizer, followed by foundation, concealer, and setting powder using beauty sponges and brushes. Eyeshadow is blended in neutral tones, eyeliner drawn precisely, and mascara applied to define the lashes. The person dots on lip tint and blush for a natural glow. The camera captures close-up details of each step. Natural light floods the scene."}

The prompt we used for Figure~\ref{fig:visible_examples_14B} is: \textit{``A fluffy domestic cat running joyfully across a sunlit meadow, its ears perked forward and tail held high with excitement. The cat's eyes are bright and focused, paws swiftly padding through the tall grass, creating natural motion blur. Golden afternoon light filters through the trees in the background, casting soft shadows. The scene radiates warmth and energy. Shot in smooth 4K slow-motion, low-angle close-up tracking shot following the cat's playful sprint."}

\clearpage
\section*{NeurIPS Paper Checklist}

\begin{enumerate}

\item {\bf Claims}
    \item[] Question: Do the main claims made in the abstract and introduction accurately reflect the paper's contributions and scope?
    \item[] Answer: \answerYes{} %
    \item[] Justification: Both the Abstract and the Introduction indicate the scope and contributions of this paper.

\item {\bf Limitations}
    \item[] Question: Does the paper discuss the limitations of the work performed by the authors?
    \item[] Answer: \answerNo{}
    \item[] Justification: The paper evaluates video diffusion models but does not include a dedicated discussion of the limitations of the proposed method.

\item {\bf Theory assumptions and proofs}
    \item[] Question: For each theoretical result, does the paper provide the full set of assumptions and a complete (and correct) proof?
    \item[] Answer: \answerYes{} %
    \item[] Justification: We provide and check the proof for theoretical results.

    \item {\bf Experimental result reproducibility}
    \item[] Question: Does the paper fully disclose all the information needed to reproduce the main experimental results of the paper to the extent that it affects the main claims and/or conclusions of the paper (regardless of whether the code and data are provided or not)?
    \item[] Answer: \answerNo{}
    \item[] Justification: We describe the algorithms and core settings, but the private fine-tuning data and incomplete training details prevent exact reproduction from the paper alone.

\item {\bf Open access to data and code}
    \item[] Question: Does the paper provide open access to the data and code, with sufficient instructions to faithfully reproduce the main experimental results, as described in supplemental material?
    \item[] Answer: \answerNo{}
    \item[] Justification: The fine-tuning dataset is private; the paper does not provide open access to all code and data needed for exact reproduction.

\item {\bf Experimental setting/details}
    \item[] Question: Does the paper specify all the training and test details (e.g., data splits, hyperparameters, how they were chosen, type of optimizer) necessary to understand the results?
    \item[] Answer: \answerNo{}
    \item[] Justification: We report model, data, sparsity, batch size, and training-step settings, but do not provide a complete optimizer and learning-rate specification.

\item {\bf Experiment statistical significance}
    \item[] Question: Does the paper report error bars suitably and correctly defined or other appropriate information about the statistical significance of the experiments?
    \item[] Answer: \answerNo{}
    \item[] Justification: We report point estimates without error bars, confidence intervals, or statistical significance tests.

\item {\bf Experiments compute resources}
    \item[] Question: For each experiment, does the paper provide sufficient information on the computer resources (type of compute workers, memory, time of execution) needed to reproduce the experiments?
    \item[] Answer: \answerNo{}
    \item[] Justification: We report the inference GPU and latencies, but not the full training hardware, memory, and compute budget for every experiment.
    
\item {\bf Code of ethics}
    \item[] Question: Does the research conducted in the paper conform, in every respect, with the NeurIPS Code of Ethics \url{https://neurips.cc/public/EthicsGuidelines}?
    \item[] Answer: \answerYes{} %
    \item[] Justification: The authors have reviewed the NeurIPS Code of Ethics.

\item {\bf Broader impacts}
    \item[] Question: Does the paper discuss both potential positive societal impacts and negative societal impacts of the work performed?
    \item[] Answer: \answerNo{}
    \item[] Justification: The paper focuses on attention efficiency and does not include a dedicated discussion of societal impacts.
    
\item {\bf Safeguards}
    \item[] Question: Does the paper describe safeguards that have been put in place for responsible release of data or models that have a high risk for misuse (e.g., pre-trained language models, image generators, or scraped datasets)?
    \item[] Answer: \answerNA{} %
    \item[] Justification: The paper does not release models, generators, or datasets with a high risk for misuse.

\item {\bf Licenses for existing assets}
    \item[] Question: Are the creators or original owners of assets (e.g., code, data, models), used in the paper, properly credited and are the license and terms of use explicitly mentioned and properly respected?
    \item[] Answer: \answerNo{}
    \item[] Justification: We cite the models and evaluation assets, but do not enumerate all asset licenses and terms of use in the paper.

\item {\bf New assets}
    \item[] Question: Are new assets introduced in the paper well documented and is the documentation provided alongside the assets?
    \item[] Answer: \answerNA{}
    \item[] Justification: The paper does not release a new dataset or model asset.

\item {\bf Crowdsourcing and research with human subjects}
    \item[] Question: For crowdsourcing experiments and research with human subjects, does the paper include the full text of instructions given to participants and screenshots, if applicable, as well as details about compensation (if any)? 
    \item[] Answer: \answerNA{} %
    \item[] Justification: The paper does not involve crowdsourcing nor research with human subjects.

\item {\bf Institutional review board (IRB) approvals or equivalent for research with human subjects}
    \item[] Question: Does the paper describe potential risks incurred by study participants, whether such risks were disclosed to the subjects, and whether Institutional Review Board (IRB) approvals (or an equivalent approval/review based on the requirements of your country or institution) were obtained?
    \item[] Answer: \answerNA{} %
    \item[] Justification: The paper does not involve crowdsourcing nor research with human subjects.

\item {\bf Declaration of LLM usage}
    \item[] Question: Does the paper describe the usage of LLMs if it is an important, original, or non-standard component of the core methods in this research? Note that if the LLM is used only for writing, editing, or formatting purposes and does \emph{not} impact the core methodology, scientific rigor, or originality of the research, declaration is not required.
    \item[] Answer: \answerNA{} %
    \item[] Justification: The core method development in this research does not involve LLMs as any important, original, or non-standard components.

\end{enumerate}

\end{document}